\documentclass{article}

\PassOptionsToPackage{numbers, compress}{natbib}

\usepackage[preprint]{neurips_2026}

\usepackage[utf8]{inputenc}
\usepackage[T1]{fontenc}
\usepackage{url}            
\usepackage{booktabs}       
\usepackage{nicefrac}       
\usepackage{xcolor}         

\usepackage{amsmath,amssymb,amsfonts,amsthm}

\usepackage{graphicx}
\usepackage{multirow}
\usepackage[protrusion=false,expansion=false]{microtype} 
\usepackage{physics}

\usepackage{algorithm}
\usepackage{algpseudocode}
\usepackage{float}
\usepackage{hyperref}       

\title{Geometric Mean Pooling for Equal-Weight Multiplicative Coarse-Graining}

\author{Ang-Kun Wu\\
Department of Physics and Astronomy, University of Tennessee, Knoxville\\
Knoxville, Tennessee 37996, USA\\
\texttt{angkunwu@gmail.com}
\And Fangdi Wen\\
Department of Physics and Astronomy, Rutgers University\\
New Brunswick, New Jersey 08901, USA\\
\And Jingtao Zhang\\
Google\\
Mountain View, CA 94043, USA\\
}
\date{}

\begin{document}

\maketitle

\begin{abstract}
As an alternative to the additive and extremal biases of average and max pooling, we introduce Geometric Mean Pooling (GMP), a signed pooling operator that combines the product of feature signs with the geometric mean of feature magnitudes. Motivated by local-to-global composition in quantum many-body physics, GMP retains both joint sign information and a characteristic multiplicative scale without introducing learnable pooling parameters. We show that non-overlapping hierarchical GMP preserves the corresponding global multiplicative statistic and evaluate it on synthetic sequence tasks, iterative coarse-graining, image classification, and molecular lipophilicity regression. On the synthetic tasks, GMP recovers product-based signals more accurately than average and max pooling and maintains predictive performance under the tested levels of multiplicative input noise. On image and molecular data, however, its effectiveness depends on the representation, target parameterization, and placement of local and global pooling. These results position GMP as a complementary, regime-dependent inductive bias for tasks in which equal-weight multiplicative composition is plausible, rather than as a universal replacement for standard pooling operators.
\end{abstract}

\section{Introduction}

Pooling is a fundamental form of coarse-graining in convolutional neural networks (CNNs): it reduces spatial or sequence resolution while retaining features useful for downstream prediction. Classical CNN architectures established local max pooling as a source of spatial selectivity and translation tolerance \citep{lecun1998gradient,krizhevsky2012imagenet}, while average pooling provides a complementary aggregation that preserves local mean responses. Global average pooling later became a standard way to replace fully connected layers with a spatially aggregated representation, reducing parameter count and encouraging correspondence between feature maps and class-level predictions \citep{lin2014network,he2016deep}. Modern surveys and learnable pooling methods have extended this basic design space to generalized means, learned norms, attention-based aggregation, and higher-order feature interactions \citep{tao2022pooling,gulcehre2013,cui2017kernel}. These operators are effective when the target is controlled by additive statistics, salient responses, or learned weighted combinations of local features. They are less naturally matched to signals encoded in the joint product of many comparable features.

Multiplicative interactions provide a distinct alternative to additive aggregation. Product units and multiplicative neural networks were introduced to represent interactions that cannot be expressed efficiently by purely additive units \citep{durbin1989product,miikkulainen1996subsymbolic}, and multiplicative gates remain central to recurrent and attention-based architectures \citep{hochreiter1997long,vaswani2017attention}. Similar product-like structure appears in scientific settings in which a global quantity is assembled from local factors, including molecular properties, variational wavefunctions, and statistical-mechanical order parameters. In the present work, we focus on the more specific equal-weight case in which the target depends on 
$
\prod_{i=1}^{N} x_i,
$
or, for positive magnitudes, equivalently on the sum $\sum_i \log |x_i|$. The individual feature marginals may then be identical across classes even though their joint products differ. In such a setting, average and max pooling can discard the relevant dependence structure, whereas a multiplicative pooling operator can preserve the product of feature signs and the average of their log-magnitudes.

Theoretically, quantum many-body states provide additional physical motivation for studying multiplicative composition. In the non-interacting limit, single-particle eigenmodes provide independent degrees of freedom, and a many-body occupation state is generated by applying creation operators to the vacuum; for example, $\ket{\Psi_{k_1,k_2}}=a^\dagger_{k_1}a^\dagger_{k_2}\ket{0}$, where $a^\dagger_k$ creates a particle in eigenstate $k$. Thus, the occupation-number basis forms a product basis in Fock space, whereas interactions mix these configurations and generate superpositions and entanglement \citep{fetter1971quantum}. For a fermionic occupation state built from single-particle orbitals, the coordinate-space wavefunction is a Slater determinant: an antisymmetrized sum of products of one-particle orbitals~\citep{slater1929theory,dirac1929theory,foulkes2001quantum}. In neural-network quantum-state representations, this determinant is commonly retained as an explicit antisymmetric component, while neural networks parameterize the orbitals and correlation factors\citep{slater1929theory,dirac1929theory,pfau2020abinitio,hermann2020abinitio}. Valence-bond-solid states provide a related example of structured multiplicative composition, with amplitudes organized from local singlet-bond factors and a superposition over compatible bond configurations \citep{anderson1973resonating,affleck1987rigorous,wuzhang2026vbs}.



With the observations mentioned above, GMP, as a parameter-free operator that averages feature log-magnitudes while retaining sign parity, provides a multiplicative complement to average and max pooling. We study global GMP, which summarizes full feature maps or sequences, and local GMP, which preserves blockwise multiplicative summaries for subsequent processing. GMP outperforms average and max pooling applied directly to inputs on synthetic signed-product classification tasks; image classification and molecular regression examine its dependence on learned representations, pooling placement, and target parameterization. Together, these studies connect the preservation of multiplicative observables across scales to pooling choice as a structural prior.

\paragraph*{Contributions}
\textbf{(1) A signed multiplicative pooling primitive.} We define a pooling operator that combines sign parity with an equal-weight geometric mean of magnitudes for local and global aggregation, and distinguish it from unsigned geometric pooling and learned arithmetic modules.
\textbf{(2) A coarse-graining analysis.} We establish hierarchical consistency for complete, equal-sized, non-overlapping groups with a shared clamp, and characterize how input signs, magnitudes, and activation choices affect the operator.
\textbf{(3) Controlled evaluation of applicability.} We organize synthetic tasks around the statistics preserved by GMP, and use image and molecular experiments to examine its dependence on representation, activation, and pooling placement. Matched comparisons with average and max pooling isolate the role of the pooling primitive.

\section{Related Work}

\paragraph{Generalized means and pooling.}
Pooling methods include fixed average and max operators, learned norms, generalized means, and higher-order feature summaries~\citep{tao2022pooling,gulcehre2013,cui2017kernel}. Generalized Mean pooling (GeM), used in image retrieval~\citep{radenovic2018fine,li2016robust}, is defined for positive inputs by
\[
\operatorname{GeM}_{p}(\mathbf{x})=
\left(\frac{1}{n}\sum_{i=1}^{n}x_i^p\right)^{1/p}.
\]
Its limit as $p\to0$ is the geometric mean. Weighted geometric pooling~\citep{weng2013learning}, alpha-integration pooling~\citep{eom2018alpha}, attention-aware GeM~\citep{gu2018attention}, and groupwise GeM~\citep{ko2022group} further explore this family.

GenAgg~\citep{kortvelesy2023genagg} provides a directly relevant framework based on generalized $f$-means, $f^{-1}(n^{-1}\sum_i f(x_i))$, with additional parameters extending the aggregation family. Its geometric-magnitude construction uses log-magnitudes. GMP builds on this established connection between geometric magnitudes and averaging in transformed coordinates. Its magnitude component inherits the equal-sized hierarchical consistency of quasi-arithmetic means. GMP fixes the magnitude transform and adds the product of input signs; for strictly positive inputs, GMP clamps the inputs from below, averages their logarithms, and exponentiates the result.

\paragraph{Learned multiplication and set representations.}
Neural Arithmetic Units include the Neural Multiplication Unit (NMU), which learns products over subsets of inputs~\citep{madsen2020neural}. Neural Power Units (NPU) learn power functions with a treatment of negative inputs through complex arithmetic~\citep{heim2020neural}. These methods address learnable arithmetic structure, whereas GMP applies a prescribed parity rule and equal log-magnitude weights. GMP's sign-parity convention provides a specific real-valued composition rule alongside these learned power functions. Deep Sets~\citep{zaheer2017deep} studies invariant representations built from learned elementwise transformations and sum aggregation. This separates two complementary design choices: selecting the aggregation statistic and learning the representation supplied to it. Our experiments focus on the first choice through matched pooling substitutions.

\paragraph{Geometry and physical coarse-graining.}
Geometric pooling on graphs and meshes concerns the geometry of the domain~\citep{xu2024geometric,milano2020mesh,bianchi2025torch,seong2018gcnn}; geometric network renormalization likewise studies structural coarse-graining~\citep{commpys2024}. Here ``geometric'' refers to a mean of feature magnitudes. Connections between neural pooling, renormalization, and tensor networks motivate a block-coarse-graining interpretation~\citep{wang2017rg,Kadanoff1966,Wilson1975,hallam2018mera,tang2016scale,PhysRevB.111.035119}. We use this interpretation to study which statistics are preserved as local summaries are composed across scales.

\section{Methodology}

\subsection{The Geometric Mean Pooling Primitive}

We define the signed geometric mean pooling (GMP) operator over a window
$W=\{x_1,\ldots,x_k\}$ as
\begin{equation}
    \mathrm{GMP}(W)
    =
    \left(\prod_{i=1}^{k}\mathrm{sign}(x_i)\right)
    \exp\!\left(
        \frac{1}{k}\sum_{i=1}^{k}\log\!\big(\max(|x_i|,\varepsilon)\big)
    \right),
    \label{eq:gp}
\end{equation}
where $\varepsilon>0$ is a lower bound on magnitudes inside the logarithm and $\operatorname{sign}(0)=0$. The implementation defaults are $10^{-6}$ in 1D and $10^{-12}$ in 2D; the hierarchical statement below assumes the same value at every level. The first factor records sign parity for nonzero inputs, and the second computes a clamped geometric magnitude. Log-space evaluation avoids forming the full magnitude product. Appendix~\ref{app:details} characterizes gradients and behavior near zero.

For nonzero inputs with an inactive clamp, GMP is related to the product decomposition
\[
\prod_{i=1}^{k}x_i
=
\left(\prod_{i=1}^{k}\mathrm{sign}(x_i)\right)
\exp\!\left(\sum_{i=1}^{k}\log |x_i|\right).
\]
Thus, GMP preserves the joint sign and an equal-weight multiplicative scale. For known $k$ and an inactive clamp, the product is recovered as $\operatorname{sign}(\mathrm{GMP})|\mathrm{GMP}|^k$. The normalization gives GMP a characteristic magnitude scale; reconstruction of the full product additionally uses the window size.

In the global regime, the window is the entire sequence, so $k=N$, and GMP reduces to a single equal-weight multiplicative summary of the whole input. In the local regime, GMP is applied blockwise with non-overlapping windows of size $k$ and stride $k$:
\begin{equation}
    \mathrm{GMP}_{\mathrm{local}}(\mathbf{x})
    =
    \left[
        \mathrm{GMP}(W_1),\ldots,\mathrm{GMP}(W_m)
    \right],
    \qquad
    m=\left\lfloor\frac{N}{k}\right\rfloor .
\end{equation}
Each complete block is replaced by its signed multiplicative summary. If $k$ does not divide $N$, this definition omits the trailing entries; the hierarchical identity applies only when the blocks cover all inputs. 

For 2D inputs, GMP extends directly to spatial pooling windows. Given a window  $W_{u,v}$ of height $r$ and width $s$, the 2D operator is
\begin{equation}
    \mathrm{GMP}_{2D}(W_{u,v})
    =
    \left(
        \prod_{h=1}^{r}\prod_{w=1}^{s}\mathrm{sign}(x_{h,w})
    \right)
    \exp\!\left(
        \frac{1}{rs}\sum_{h=1}^{r}\sum_{w=1}^{s}\log\!\big(\max(|x_{h,w}|,\varepsilon)\big)
    \right),
\end{equation}
and is applied independently to each channel unless otherwise specified.

\subsubsection*{Renormalization-group perspective and algebraic contrast}

From a machine-learning perspective, a pooling layer replaces each local group of activations with a lower-resolution feature. Repeated application therefore composes local summaries into progressively coarser representations, which can be formalized as a coarse-graining map: each block is replaced by a single summary variable. 
In the language of Kadanoff block-spin RG \citep{Kadanoff1966,Wilson1975}, a lattice of microscopic variables $\{x_i\}_{i=1}^N$ is partitioned into non-overlapping blocks $B_j$ of size $b$, and each block is replaced by a single block variable $x'_j=\mathcal R_b(B_j)$. Repeated application generates a flow
\[
x^{(0)} \xrightarrow{\ \mathcal R_b\ } x^{(1)} \xrightarrow{\ \mathcal R_b\ } x^{(2)}
\xrightarrow{\ \mathcal R_b\ } \cdots \xrightarrow{\ \mathcal R_b\ } x^{(n)},
\qquad N_n = N / b^n.
\]
An order parameter is a fixed point of this flow if its value is preserved across scales; in other words, repeated blockwise coarse-graining yields the same statistic as a single global summary. This is the natural RG language for distinguishing pooling operators.

For average pooling, the block map is the additive block-spin transform
$
\mathcal R_b^{\mathrm{avg}}(B)=\frac{1}{b}\sum_{i\in B}x_i,
$
and for max pooling it is the extremal order-statistic map
$
\mathcal R_b^{\max}(B)=\max_{i\in B}x_i.
$
These are natural RG maps for linear or extremal observables, but they are not fixed points for equal-weight multiplicative statistics. By contrast, GMP defines the block map
$
\mathcal R_b^{\mathrm{GMP}}(B)
=
\left(\prod_{i\in B}\mathrm{sign}(x_i)\right)
\exp\!\left(
\frac{1}{b}\sum_{i\in B}\log\!\big(\max(|x_i|,\varepsilon)\big)
\right).
$
Under the change of variables $u_i=\log\!\big(\max(|x_i|,\varepsilon)\big)$, the magnitude part of GMP becomes an arithmetic average in log-magnitude space:
\[
\log |\mathcal R_b^{\mathrm{GMP}}(B)|=\frac{1}{b}\sum_{i\in B}u_i.
\]
Hence GMP is algebraically conjugate to additive averaging in log-space, and repeated non-overlapping block pooling preserves the global multiplicative statistic exactly (proof in Appendix~\ref{app:hierarchical}). In this precise sense, GMP is a fixed-point RG map for equal-weight multiplicative order parameters.

When the target depends on a product of many features, average and max pooling may discard information needed to recover the target. In Case A of the experiments below, GMP directly retains the sign parity that determines the class label, whereas the tested average- and max-pooling classifiers perform near chance in the independent-input setting.

\subsection{Local and Global Placement}

Global GMP summarizes a full sequence or feature map and is structurally aligned with a global equal-weight multiplicative statistic. Local GMP preserves blockwise statistics for downstream processing and is aligned with tasks in which the relevant factors occur within those blocks. These two placements let the pooling scale reflect the spatial or sequential organization of the signal.

The pooling inputs determine the role of the sign component. With strictly positive inputs, GMP equals the geometric mean of clamped inputs. With ReLU inputs, any zero in a window forces its signed-GMP output to zero, and the implemented backward pass gives zero derivatives for all inputs in that window. Thus, activation choice affects both the information available to pooling and its optimization behavior.

\section{Experiments}

The synthetic tasks examine prescribed sign and magnitude statistics; the application experiments examine pooling within learned representations.

\subsection{Synthetic Sequence Classification}

\begin{figure*}[t!]
\centering
\includegraphics[width=0.49\textwidth]{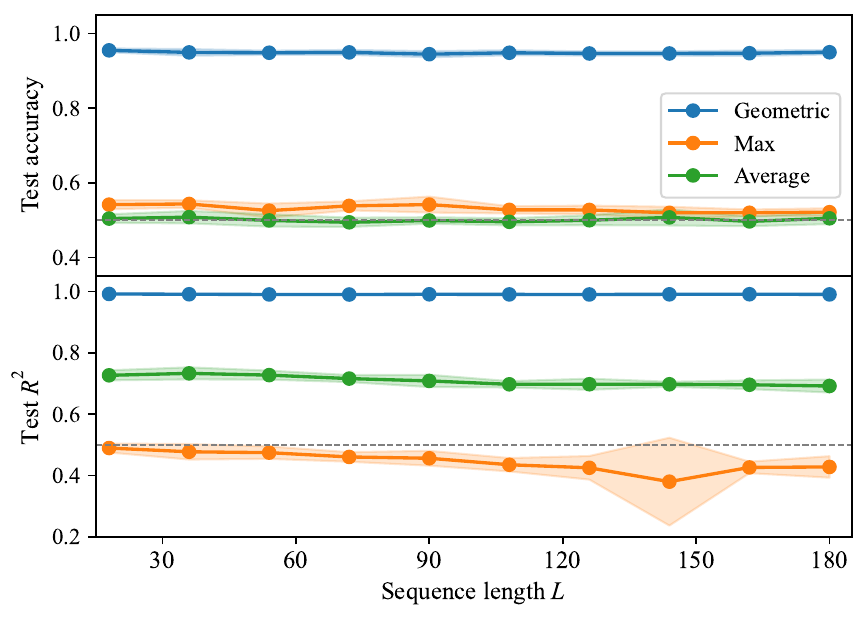}
\includegraphics[width=0.49\textwidth]{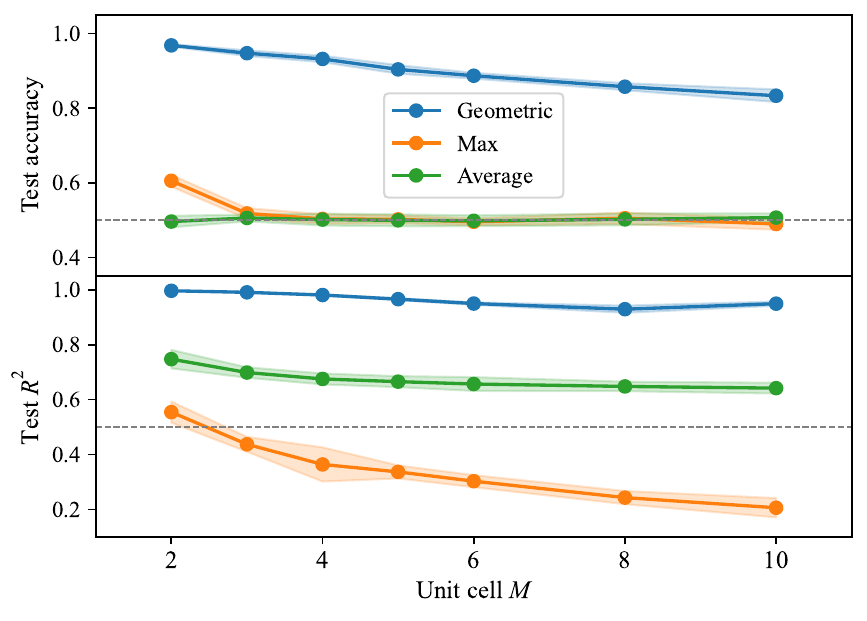}
\caption{Sequence-length and cell-size sweeps for Case B classification (top) and $R^2$ regression (bottom). Left: varying sequence length with cell size $3$. Right: varying cell size with sequence length $120$. Clean classification inputs are standard Gaussian; clean regression inputs are lognormal with standard Gaussian logarithms. Targets are computed before adding Gaussian input noise with standard deviation $0.05$. Error bars denote the standard deviation across ten random seeds.
}
\label{fig:seq_uc_scaling}
\end{figure*}


\noindent\textbf{Case A (global signed-product classification).}
For each example, we draw a clean sequence
$x_i \overset{\mathrm{i.i.d.}}{\sim} \mathcal{N}(0,1)$ of length $N$ and set the label $y = \mathbf{1}\{\prod_{i=1}^{N} x_i < 0\}$, with each $x_i$ being independent identically distributed (i.i.d.).
Equivalently, the two classes differ only in the parity of negative entries: class $0$ has an even number of negative entries and class $1$ an odd number. For this independent-input setting ($N>1$), the class prior is balanced and every individual coordinate has the same marginal distribution in both classes. We compare global max, average, and GMP, each followed by a linear classifier: $
\mathrm{Pool}_{\mathrm{global}} \;\to\; \mathrm{FC}(1,2),
$
where FC denotes a fully connected layer mapping.
No convolution, normalization, or pointwise activation is used, so that the pooling operator acts directly on the variables carrying the product structure. Across the reported single-seed sweeps over sequence length $N\in\{16,32,64,128\}$ and adjacent-site correlation
$\rho\in\{0,0.3,0.5,0.7,0.9\}$, GMP achieves $100\%$ accuracy. Average and max pooling remain near chance for independent inputs and both reach $60.4\%$ accuracy at $\rho=0.9$.
Sweep details and full results appear in Appendix~\ref{app:global_synthetic}, Table~\ref{tab:case_a}.

\noindent\textbf{Case B (local signed-product-sum classification).}
To test local multiplicative coarse-graining, we partition a clean sequence $x_i \overset{\mathrm{i.i.d.}}{\sim}\mathcal{N}(0,1)$ into non-overlapping cells $C_j$ of size $k$. Each cell contributes the signed geometric mean
\[
g_j =
\left(\prod_{i\in C_j}\operatorname{sign}(x_i)\right)
\exp\!\left(
\frac{1}{k}\sum_{i\in C_j}\log(\max\{|x_i|,\varepsilon\})
\right),
\]
and the sample-level statistic is
$
S = \sum_{j=1}^{N/k} g_j,\;
y=\mathbf{1}\{S<0\}.
$
Thus the label depends on a sum of local, equal-weight multiplicative
statistics. The classifier receives a noisy observation
$\tilde{x}_i=x_i+\eta_i$, where $\eta_i\sim\mathcal{N}(0,0.05^2)$ independently; labels are always computed from the clean sequence.

For every pooling primitive, we use the matched architecture
\[
\mathrm{Pool}_{\mathrm{local}}(k,\mathrm{stride}=k)
\;\to\; \mathrm{Flatten}
\;\to\; \mathrm{FC}(N/k,2).
\]
The local windows are therefore aligned exactly with the generative cells. As in Case A, there are no convolutional layers, normalizations, or activations before pooling. This deliberately minimal setting isolates the information preserved by the pooling primitive: on clean inputs with matching numerical conventions, GMP produces the local statistics $g_j$ directly, after which the linear classifier can learn their sum. Noisy observations test how accurately these local summaries retain the clean class signal.

The classification panels of Fig.~\ref{fig:seq_uc_scaling} show a consistent advantage for GMP over average and max pooling in the aligned settings. Accuracy remains high across the sequence-length sweep and declines as the cell size grows. The two sweeps illustrate how the operator's multiplicative prior interacts with the number and size of local groups under input noise.

\noindent\textbf{Impact of Sign Information.}
To quantify the effect of sign information on classification accuracy, we perform an ablation study comparing signed and unsigned geometric-mean pooling (Table~\ref{tab:sign_ablation}). We first embed each input sequence into a higher-dimensional space to increase model expressivity, using the architecture
\[
\mathrm{Conv1d}(1,d_{\mathrm{embed}},k=1)\;\to\;\mathrm{Pool}\;\to\;\mathrm{FC}(d_{\mathrm{embed}},2),
\]
with $d_{\mathrm{embed}}=32$. We evaluate both Case A and Case B under the corresponding global and local pooling settings. The only difference between the pooling variants is whether the geometric mean is computed with or without the sign-parity term. Removing the sign-parity term reduces accuracy to approximately chance level in both tasks, demonstrating its importance in these settings.

\begin{table}[H]
\caption{Ablation of the sign-parity term in geometric mean pooling. Results are reported as mean $\pm$ standard deviation across $10$ random seeds. The sequence length is $120$, and the cell size is $3$ for Case B.}
\label{tab:sign_ablation}
\centering
\begin{tabular}{l c}
\toprule
\textbf{Setting} & \textbf{Accuracy} \\
\midrule
Case A, signed & 0.9330 $\pm$ 0.0659 \\
Case A, unsigned & 0.5107 $\pm$ 0.0165 \\
Case B, signed & 0.9856 $\pm$ 0.0047 \\
Case B, unsigned & 0.5072 $\pm$ 0.0136 \\
\bottomrule
\end{tabular}
\end{table}

\subsection{Synthetic Magnitude Regression and Iterative Pooling}

The synthetic regression tasks use MSE loss and a linear output layer. Lognormal clean inputs provide a controlled setting for studying multiplicative magnitudes. Local aggregation, learned embeddings, and input perturbations then probe how this structure is carried into the prediction.

\noindent\textbf{R1 (Global geometric-mean regression).}
The clean task draws $x_i\sim\mathrm{lognormal}(0,1)$ independently and sets $y=\exp(N^{-1}\sum_i\log x_i)$. The model uses a learned pointwise $\mathrm{Conv1d}(1,d_{\mathrm{embed}},k=1)$, followed by variable global pooling and a linear regression head, with $d_{\mathrm{embed}}=1$ for R1. The geometric mean of the raw inputs provides the exact target statistic when the clamp is inactive. The learned affine embedding and regression head allow the model to adapt this summary during training. 

For $N\in\{16,32,64,128\}$ in the reported single-seed experiment, GMP achieves $R^2=0.993$--$1.000$, compared with $0.609$--$0.638$ for average pooling and $0.019$--$0.137$ for max pooling. These results support the correspondence between geometric aggregation and the multiplicative target; average pooling also retains predictive information through correlations between arithmetic and geometric magnitudes. Full results appear in Appendix~\ref{app:global_synthetic}, Table~\ref{tab:r1}.

\noindent\textbf{R2 (Local cell log-mean regression):} 
We use i.i.d. lognormal features, $x_i \sim \mathrm{lognormal}(0,1)$, with cell size $c \in \{2,3,4,5,6,8,10\}$ and sequence length $N \in \{18,36,54,72,90,108,126,144,162,180\}$ (fixed $c=3$) or $N=120$ (varying $c$). Each cell contributes $s_j = \exp(\frac{1}{c}\sum_{i\in \mathrm{cell}_j}\log|x_i|)$, and the target $y=\frac{1}{n_\mathrm{cell}}\sum_j s_j$ is computed from the \emph{clean} sequence. The architecture is local pooling (kernel and stride $c$), flattening, and $\mathrm{FC}(n_\mathrm{cell},32)\to\mathrm{ReLU}\to\mathrm{FC}(32,1)$. Pooling is selected from GMP, average, and max and acts directly on the noisy sequence, without a preceding convolutional embedding. The sweeps appear in Fig.~\ref{fig:seq_uc_scaling} (bottom panels). Observations are perturbed as $\tilde{x}_i = x_i + \eta_i$, with $\eta_i \sim \mathcal{N}(0, 0.05^2)$.

\noindent\textbf{R3 (Noise robustness):} Clean lognormal features are corrupted by multiplicative noise applied to every feature, followed by additive noise: $\tilde{x}=x\cdot\exp(\epsilon_\mathrm{mul})+\epsilon_\mathrm{add}$, $\epsilon_\mathrm{mul}\sim\mathcal N(0,\sigma_\mathrm{mul}^2)$, $\epsilon_\mathrm{add}\sim\mathcal N(0,\sigma_\mathrm{add}^2)$. The target is the geometric mean of the clean features. The model uses the same architecture as R1, but with $d_{\mathrm{embed}}=32$ to increase the capacity of the learned representation before global pooling.

\begin{figure}[t]
\centering
\includegraphics[width=1.0\textwidth]{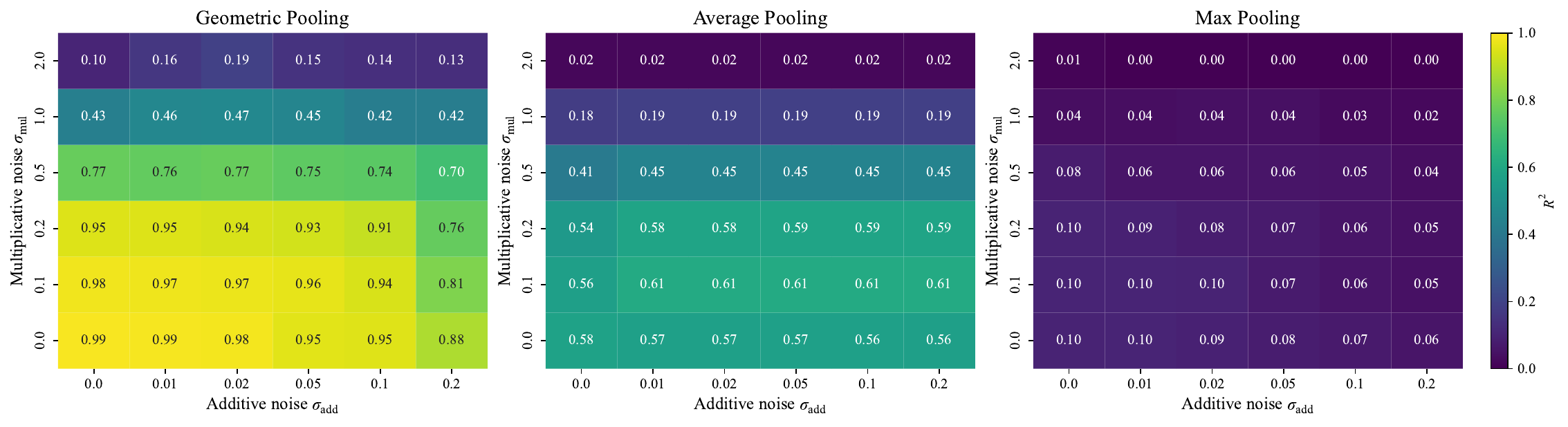}
\caption{
R3: global regression under multiplicative and additive input noise, $\tilde{x}=x\exp(\epsilon_{\mathrm{mul}})+\epsilon_{\mathrm{add}}$, with $\log x\sim\mathcal N(0,1)$ and independent Gaussian perturbations of standard deviations $\sigma_{\mathrm{mul}}$ and $\sigma_{\mathrm{add}}$. Over the tested noise ranges, GMP maintains higher predictive performance than max pooling, including at low noise levels.
}

\label{fig:noise_heatmaps}
\end{figure}

Figure~\ref{fig:noise_heatmaps} compares responses to multiplicative and additive perturbations. Multiplicative noise shifts raw log-magnitudes additively, directly matching the coordinates used by GMP. Additive noise probes a different perturbation and can also change input signs. The sweep examines these two effects within the same prediction task.

\noindent\textbf{R4 (Iterative preservation of a multiplicative statistic).}
Clean lognormal sequences of length $L=256$ are perturbed once by multiplying a randomly selected fraction $p=25\%$ of entries by $\exp(\eta)$, with $\eta\sim\mathcal N(0,1)$. Repeated pooling uses kernel and stride $2$, reducing $256$ entries to one in eight steps. At every step, the evaluator takes a global geometric-mean readout of the remaining entries for \emph{all three} pooling methods and compares it with the clean global geometric mean using MAE.

\begin{figure}[t]
\centering
\includegraphics[width=0.5\textwidth]{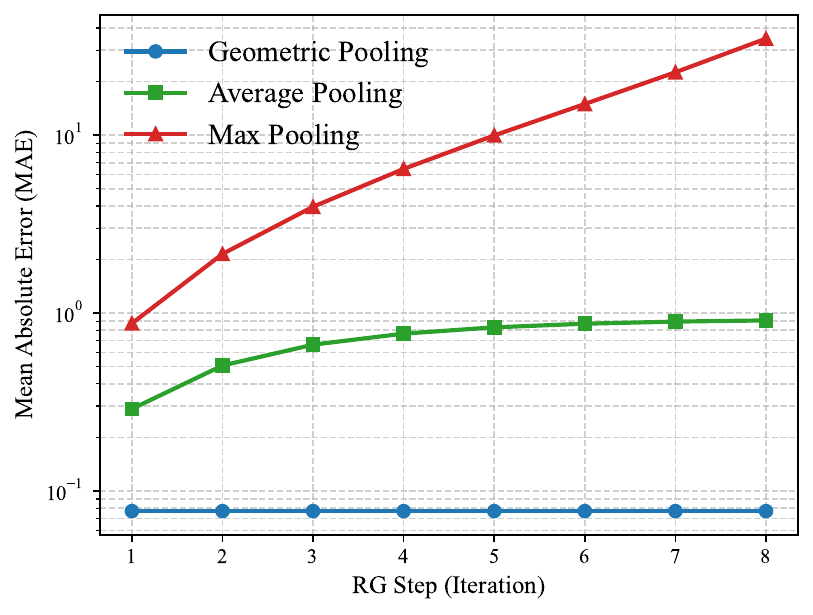}
\caption{R4: preservation of a multiplicative observable across pooling levels. MAE is evaluated against the clean geometric-mean target ($\sigma=1$, $p=25\%$, $L=256$), using a global geometric readout after each local pooling step for all methods. GMP maintains an approximately constant readout error across the hierarchy.}

\label{fig:rg_flow}
\end{figure}

Figure~\ref{fig:rg_flow} illustrates the connection between hierarchical consistency and preservation of a multiplicative observable. The GMP curve remains approximately constant across successive pooling steps, whereas average and max pooling change the common geometric readout. This behavior reflects the statistic preserved by each block transformation: GMP composes geometric summaries, average pooling composes arithmetic means, and max pooling composes extrema.

\subsection{Image Classification and Activation Compatibility}

\begin{figure}[h!]
\centering
\includegraphics[width=1.0\textwidth]{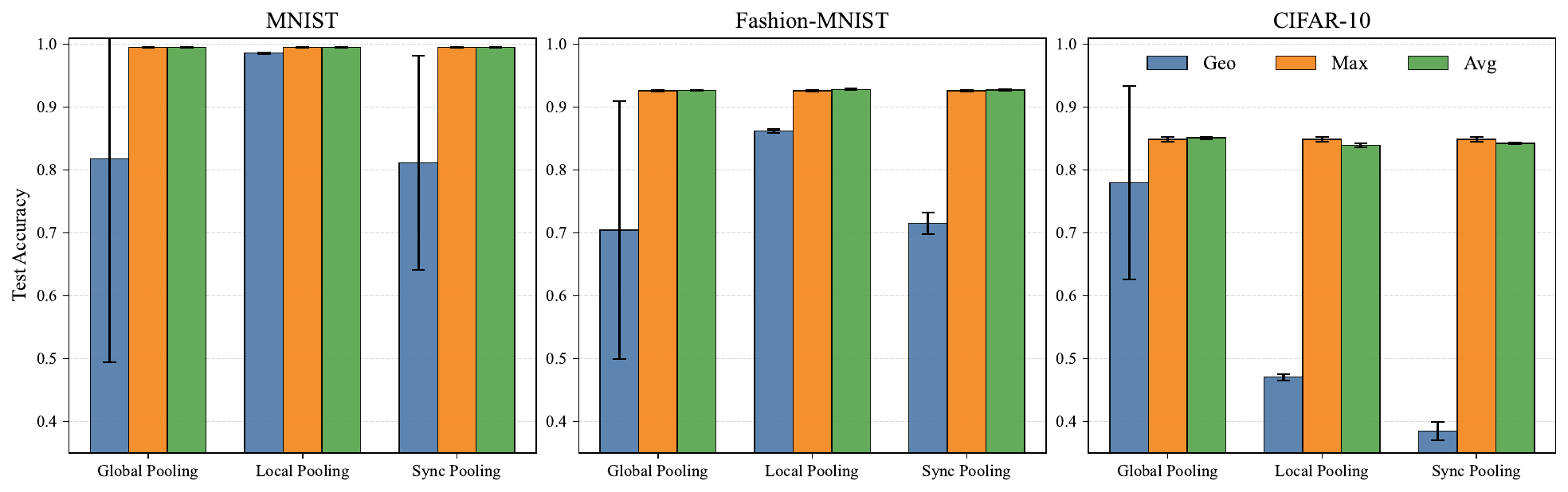}
\caption{
Validation accuracy on MNIST, Fashion-MNIST, and CIFAR-10 across ten random seeds. Each panel reports the three pooling primitives (GMP, max, and average) under three architectures: variable global pooling with fixed intermediate max pooling, variable local pooling with fixed global max pooling, and synchronized local-plus-global pooling. Local GMP is generally more competitive than global GMP on MNIST and Fashion-MNIST, whereas the gap between GMP and the gap between GMP and the average- and max-pooling baselines increases on CIFAR-10. Error bars denote standard deviation across 10 seeds. All models are trained with cross-entropy loss and Adam~\citep{kingma2015adam} using batch size 128 and learning rate $10^{-3}$.
}
\label{fig:mnist_bars}
\end{figure}

We evaluate GMP on MNIST~\citep{lecun1998gradient}, Fashion-MNIST~\citep{xiao2017fashion}, and CIFAR-10~\citep{krizhevsky2009learning} using convolutional classifiers with matched channel dimensions, optimizers, and training procedures. The experiments isolate three different placements of the variable pooling primitive: global aggregation, repeated local spatial reduction, and synchronized local-plus-global pooling. For each configuration, we compare GMP, max pooling, and average pooling.

The three configurations specify where the variable pooling operator is
used:
\[
\begin{array}{c|cc}
\textbf{Configuration} & \textbf{Local pooling} &
\textbf{Global pooling} \\
\hline
\text{Global} & \text{fixed max} & \text{variable} \\
\text{Local} & \text{variable} & \text{fixed max} \\
\text{Synchronized} & \text{variable} & \text{same as local}
\end{array}
\]
The variable operator is selected from $\{\mathrm{GMP},\mathrm{Max},\mathrm{Avg}\}$. Thus, the Global configuration isolates the final spatial summary, the Local configuration tests repeated local coarse-graining, and the Synchronized configuration applies the same pooling operator at both levels. All backbones use convolution--ReLU blocks, with matched architectures and training procedures within each dataset. Full architecture details appear in Appendix~\ref{app:mnist}.

Figure~\ref{fig:mnist_bars} shows that local GMP remains competitive with max and average pooling on MNIST and Fashion-MNIST, whereas global GMP performs less consistently across the tested settings. The performance gap is larger on CIFAR-10. The sensitivity of signed GMP to exact zeros after ReLU offers a possible explanation for the weaker global results. The results therefore support a regime-specific interpretation of GMP: it is a useful structural prior when multiplicative composition is plausible, but it is not generally interchangeable with additive pooling for generic image classification.

\subsection{Lipophilicity Regression}

We evaluate on the Lipophilicity regression task from the MoleculeNet molecular benchmark suite~\citep{ramsundar2017}, specifically version 1 (V1), curated from ChEMBL with 4,200 small organic molecules. Molecules are partitioned by a Bemis--Murcko scaffold split~\citep{bemis1996properties}, with approximately 80\% of scaffolds assigned to training and the remaining scaffolds assigned to testing, so the test set contains molecular scaffolds absent from training. The dataset label is a log-scale lipophilicity value, $y=\log D$, the experimental octanol/water distribution coefficient at pH~7.4 (range approximately $[-1.5, 4.5]$). To restore the true multiplicative structure of the molecular property, we transform the regression target as $z=\exp(y)$ before optimization and evaluation, so the reported RMSE and $R^2$ are measured on the $z$ scale (range approximately $[0.2, 90]$) rather than the $\log D$ scale.

Each molecule is represented by a 2,048-bit Morgan fingerprint (Morgan FP) with radius 2 (ECFP4)~\citep{rogers2010extended}. Because the raw fingerprint is fixed, high-dimensional, and sparse, our 2D CNN maps it through a fully connected projection layer, $\mathbb{R}^{2048}\to\mathbb{R}^{1024}$ followed by ReLU, giving the network learnable capacity to re-express the fingerprint into a denser representation rather than operating directly on raw binary bits. The projected vector is reshaped into a single-channel $32\times32$ latent feature map, to which a softplus activation is applied before three convolutional blocks, each consisting of a $3\times3$ convolution (channel widths $1\to32\to64\to128$), the same activation, and a $2\times2$ local pooling operator with stride 2 (reducing the spatial resolution $32\times32\to16\times16\to8\times8\to4\times4$). The resulting $128$-channel feature map is reduced by a global pooling operator to a $128$-dimensional vector, which is passed to a two-layer regression head, $\mathrm{FC}(128,32)\to\mathrm{ReLU}\to\mathrm{FC}(32,1)$. For each experiment, the local and global pooling operators are independently selected from geometric, max, and average pooling. The learned $32\times32$ layout supplies latent neighborhoods for convolutional processing, with locality defined by the representation.


We compare the 2D CNN against four baselines trained on the same scaffold split, summarized in Table~\ref{tab:lipo_baselines}: Morgan fingerprint features with XGBoost~\citep{chen2016xgboost}, a Random Forest regressor~\citep{breiman2001random}, and a fully connected network, as well as a learned SMILES~\citep{weininger1988smiles} token embedding passed through the same fully connected head. The three Morgan-fingerprint baselines achieve similar performance and outperform the tested SMILES-embedding baseline, suggesting that Morgan fingerprints provide a more effective representation in this experimental setting.

\begin{table}[htbp]
\caption{Lipophilicity regression performance on the original logarithmic label scale, using a scaffold split.
}
\label{tab:lipo_baselines}
\centering
\begin{tabular}{l|cc}
\hline
\textbf{Baseline} & \textbf{RMSE} & \textbf{$R^2$} \\ \hline
Morgan FP + XGBoost      & 0.8870 & 0.4631 \\
Morgan FP + Random Forest & 0.8718 & 0.4813 \\
Morgan FP + FC           & 0.8752 & 0.4773 \\
SMILES Embedding + FC    & 1.1596 & 0.0824 \\ \hline
\end{tabular}
\end{table}

\begin{figure}[t]
\centering
\includegraphics[width=0.6\textwidth]{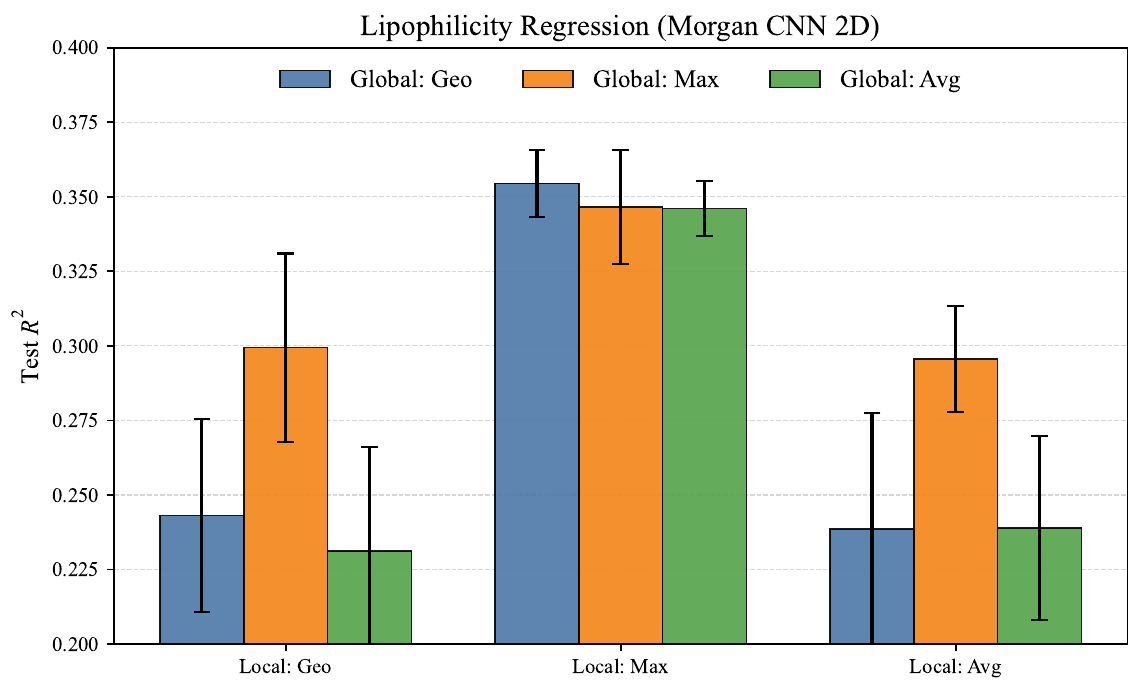}
\caption{
Test $R^2$ on the auxiliary target $z=\exp(y)$ for the Morgan-fingerprint 2D CNN under all combinations of local and global pooling operators. Each molecule is represented by a radius-2, 2,048-bit Morgan fingerprint, projected to a learned $32\times32$ latent feature map, and processed by a three-block convolutional hierarchy. Values and error bars denote the mean and standard deviation over ten random seeds.
}

\label{fig:lipo}
\end{figure}

Fig.~\ref{fig:lipo} shows the $R^2$ performance of the Morgan FP 2D CNN under all combinations of local and global pooling operators. Global GMP outperforms max and average pooling, while local GMP becomes detrimental when combined with global max pooling.  The best pooling configuration achieves $R^2=0.3545$ on the auxiliary target $z$. Because the baselines report results on the log scale of $z$, those values provide an indirect comparison with this experiment. The results overall highlight the importance of matching the pooling strategy to the underlying algebraic structure of the target property. As shown in Appendix~\ref{app:lipo}, training the same architecture on the original label scale changes the ranking, with max pooling achieving the strongest performance.

\section{Conclusion}

We introduced GMP, a parameter-free pooling operator that combines sign parity with equal-weight geometric magnitudes. It preserves the corresponding global statistic under complete, equal-sized, non-overlapping hierarchical pooling with a shared clamp and no intervening transformations.
Synthetic experiments demonstrate its utility for prescribed multiplicative signals, while image and molecular experiments show that its effectiveness depends on representation, activation, pooling placement, and target parameterization. GMP therefore provides a structural prior for sign-sensitive or equal-weight multiplicative tasks. For positive targets,
predicting their logarithms with additive aggregation offers an alternative. Extending this framework to complex-valued representations \citep{trabelsi2018deep}, developing learnable hybrid operators, and testing them on physical~\citep{wu2025modeling,carleo2017solving,pfau2020abinitio}, chemical~\citep{gilmer2017neural,schutt2017schnet,yang2019analysing}, and other scientific datasets are important directions for future work.

\begin{ack}
The work at UT Knoxville was primarily supported by the National Science Foundation Materials Research Science and Engineering Center program through the UT Knoxville Center for Advanced Materials and Manufacturing (DMR-2309083). Computations were performed using the University of Tennessee Infrastructure for Scientific Applications and Advanced Computing (ISAAC) computational resources. All experiments used an NVIDIA RTX A6000 GPU.
\end{ack}

\subsection*{Reproducibility statement}

The results presented in this paper are readily reproducible based on the descriptions, algorithms, and code provided in the main text, appendix. The codes are publicly available at \url{https://github.com/angkun-research/GeometricMeanPooling}.

\bibliographystyle{plainnat}
\bibliography{references}

\newpage
\appendix

\section{Implementation details and pseudocode}\label{app:details}

GMP operates independently on each batch item and channel.  In local mode, it reduces each non-overlapping (or strided) window; in global mode, the window is the full remaining spatial domain.  For nonzero inputs, the signed output retains the parity of negative values, while the magnitude is evaluated in log-space to avoid forming the full product.

\begin{algorithm}[H]
\caption{Signed Geometric Mean Pooling (GMP)}
\label{alg:gmp}
\begin{algorithmic}[1]
\Require Input tensor $X$ of shape $(B,C,L)$ in 1D or $(B,C,H,W)$ in 2D;
window size $K$; stride $S$; magnitude lower bound $\varepsilon$
\Ensure Pooled tensor $Y$

\If{global pooling is selected}
    \State Let $\mathcal{W}$ be the full sequence (1D) or full spatial map (2D)
\Else
    \State Extract sliding windows $\{\mathcal{W}_j\}$ of size $K$ and stride $S$
    \Comment{$S=K$ by default}
\EndIf

\For{each batch item $b$, channel $c$, and window $\mathcal{W}_j$}
    \State $s \gets \prod_{x \in \mathcal{W}_j} \operatorname{sign}(x)$
    \State $\ell \gets \frac{1}{|\mathcal{W}_j|}
        \sum_{x \in \mathcal{W}_j} \log\!\big(\max(|x|,\varepsilon)\big)$
    \State $Y_{b,c,j} \gets s \cdot \exp(\ell)$
\EndFor

\State \Return $Y$
\end{algorithmic}
\end{algorithm}

For 1D local pooling, $\mathcal{W}_j$ contains $k$ sequence entries and the output has shape $(B,C,L_{\mathrm{out}})$, where $L_{\mathrm{out}}=1+\lfloor(L-k)/s\rfloor$. For 2D local pooling, each window contains $k_hk_w$ spatial entries and the output has shape $(B,C,H_{\mathrm{out}},W_{\mathrm{out}})$. Global pooling returns
$(B,C,1)$ in 1D and $(B,C,1,1)$ in 2D.

The implementation uses \verb|torch.sign|, whose autograd derivative is zero, and a clamped log-magnitude computation. For a window of nonzero inputs, away from $|x_j|=\varepsilon$, its derivative is
\[
\frac{\partial G_\varepsilon}{\partial x_j}=
\begin{cases}
G_\varepsilon(\mathbf{x})/(k x_j), & |x_j|>\varepsilon,\\
0, & 0<|x_j|<\varepsilon.
\end{cases}
\]
The sign is locally constant within each orthant; learning there proceeds through magnitudes. Small inputs above the clamp can amplify derivatives relative to the output scale, whereas inputs below it have zero magnitude derivatives. If any input equals zero, the sign product vanishes and the implemented backward pass yields zero derivatives for \emph{every} input in that window, including nonzero entries.

Clamping also makes the signed operator discontinuous at zero. With $k-1$ inputs equal to $1$ and $0<\varepsilon<1$, the remaining input $t$ gives
\[
G_\varepsilon(t,1,\ldots,1)=
\begin{cases}
\operatorname{sign}(t)\,\varepsilon^{1/k}, & 0<|t|<\varepsilon,\\
0, & t=0.
\end{cases}
\]
Thus arbitrarily small sign changes can produce a finite output jump. ReLU removes negative signs but introduces exact zeros; Softplus gives positive values in exact arithmetic and removes the negative-parity mechanism.

\subsection{Image-classification architectures}

\section{Proof: hierarchical consistency of signed GMP}
\label{app:hierarchical}

Let $x_1,\dots,x_N$ be the inputs and fix $\varepsilon>0$. Partition the indices into $m$ non-overlapping blocks of size $k$, so that $N=mk$. Define
\[
G_\varepsilon(B)
=
\left(\prod_{i\in B}\operatorname{sign}(x_i)\right)
\exp\left(
\frac{1}{|B|}
\sum_{i\in B}
\log\!\big(\max(|x_i|,\varepsilon)\big)
\right).
\]

If any block contains a zero, then its sign product is zero and its GMP output is zero. Consequently, the sign product of the second-level GMP is also zero, which agrees with the direct GMP because the original
input contains a zero.

It remains to consider the case in which all inputs are nonzero. For each block $B_j$, every magnitude $\max(|x_i|,\varepsilon)$ is at least $\varepsilon$, and hence
\[
|G_\varepsilon(B_j)|\geq\varepsilon.
\]
Therefore, the outer clamp is inactive:
\[
\max\big(|G_\varepsilon(B_j)|,\varepsilon\big)
=
|G_\varepsilon(B_j)|.
\]
Thus,
\begin{align*}
&G_\varepsilon\big(
G_\varepsilon(B_1),\ldots,G_\varepsilon(B_m)
\big)\\
&=
\left(
\prod_{j=1}^{m}
\prod_{i\in B_j}\operatorname{sign}(x_i)
\right)
\exp\left[
\frac{1}{m}
\sum_{j=1}^{m}
\frac{1}{k}
\sum_{i\in B_j}
\log\!\big(\max(|x_i|,\varepsilon)\big)
\right]\\
&=
\left(\prod_{i=1}^{N}\operatorname{sign}(x_i)\right)
\exp\left[
\frac{1}{N}
\sum_{i=1}^{N}
\log\!\big(\max(|x_i|,\varepsilon)\big)
\right]\\
&=
G_\varepsilon(x_1,\ldots,x_N).
\end{align*}

Hence the identity holds in exact arithmetic for the clamped definition, including zero-valued inputs, when equal-sized, non-overlapping blocks cover every input, the same $\varepsilon$ is used at both levels, and $\operatorname{sign}(0)=0$. It extends inductively to such hierarchies without intervening feature transformations. Floating-point implementations may differ by rounding.

\section{Additional results for global synthetic tasks}
\label{app:global_synthetic}

This section reports the complete results for Case A and R1, which test global pooling on prescribed multiplicative targets. Both tables report single-seed experiments using seed $42$.

\subsection{Case A: Global signed-product classification}

Case A predicts the sign parity of a sequence using global pooling followed by a linear classifier, without a preceding convolution, normalization, or pointwise activation.
We vary the sequence length over $N\in\{16,32,64,128\}$ with independent standard Gaussian inputs. A separate sweep fixes $N=32$ and varies the adjacent-site correlation over $\rho\in\{0,0.3,0.5,0.7,0.9\}$. Correlated sequences are generated recursively as
\[
x_1\sim\mathcal{N}(0,1), \qquad
x_i=\rho x_{i-1}+\sqrt{1-\rho^2}\,z_i,
\]
where $z_i\sim\mathcal{N}(0,1)$ are independent innovations. The label is determined by the sign of the resulting sequence product.

Table~\ref{tab:case_a} shows that GMP achieves perfect accuracy in all reported settings. Average and max pooling remain near chance for independent inputs, while both reach $0.6040$
accuracy at $\rho=0.9$.

\begin{table}[htbp]
\caption{
Case A classification accuracy with seed $42$. The correlation sweep fixes $N=32$; the sequence-length sweep uses independent inputs ($\rho=0$).
}
\label{tab:case_a}
\centering
\begin{tabular}{l c c c c}
\toprule
\textbf{Sweep} & \textbf{Value} & \textbf{Max acc} & \textbf{Average acc} & \textbf{GMP acc} \\
\midrule
\multirow{5}{*}{$\rho$-sweep (seed=42, seq\_len=32)} 
  & 0.0   & 0.4680 & 0.5390 & 1.0000 \\
  & 0.3   & 0.5080 & 0.4860 & 1.0000 \\
  & 0.5   & 0.5000 & 0.4900 & 1.0000 \\
  & 0.7   & 0.5130 & 0.4830 & 1.0000 \\
  & 0.9   & 0.6040 & 0.6040 & 1.0000 \\
\addlinespace
\multirow{4}{*}{Seq\_len sweep (seed=42, $\rho=0$)} 
  & 16    & 0.5190 & 0.5190 & 1.0000 \\
  & 32    & 0.4680 & 0.5390 & 1.0000 \\
  & 64    & 0.4860 & 0.4710 & 1.0000 \\
  & 128   & 0.5050 & 0.5050 & 1.0000 \\
\bottomrule
\end{tabular}
\end{table}

\subsection{R1: Global geometric-mean regression}

R1 predicts the geometric mean of independent $\mathrm{lognormal}(0,1)$ inputs. The model applies a learned pointwise convolution with $d_{\mathrm{embed}}=1$, global pooling, and a linear regression head. Inputs are noiseless, and the sequence lengths are $N\in\{16,32,64,128\}$.

Table~\ref{tab:r1} reports both $R^2$ and MSE. GMP achieves $R^2$ between $0.993$ and $1.000$, whereas average pooling achieves $0.609$--$0.638$ and max pooling achieves $0.019$--$0.137$. This experiment tests geometric aggregation on an explicitly matched target.

\begin{table}[htbp]
\caption{
R1 global geometric-mean regression on noiseless lognormal inputs with seed $42$. Results are reported as $R^2$ and MSE; MSE values are scaled by $10^3$.
}
\label{tab:r1}
\centering
\begin{tabular}{c|ccc|ccc}
\hline
\multirow{2}{*}{\textbf{$N$}} & \multicolumn{3}{c|}{$R^2$} & \multicolumn{3}{c}{MSE ($\times 10^{-3}$)} \\ \cline{2-7}
& Max & Avg & GMP & Max & Avg & GMP \\ \hline
16 & 0.137 & 0.609 & 1.000 & 60.7 & 27.5 & 0.0 \\
32 & 0.081 & 0.620 & 1.000 & 29.3 & 12.1 & 0.0 \\
64 & 0.071 & 0.631 & 0.993 & 14.6 & 5.8 & 0.0 \\
128 & 0.019 & 0.638 & 1.000 &  7.9 & 2.9 & 0.0 \\ \hline
\end{tabular}
\end{table}

\section{Image-classification architectures}\label{app:mnist}

\paragraph{CNN architectures.}
We use convolutional classifiers in which the pooling operators are the only experimental variables. For MNIST and Fashion-MNIST, the backbone contains three convolution--ReLU--pooling blocks with 32 channels, followed by a fourth convolution--ReLU block. Each convolution uses a $3\times3$ kernel with padding one. Local pooling uses a non-overlapping $2\times2$ window with stride two, reducing the spatial resolution from $28\times28$ to $3\times3$. The resulting 32-channel representation is globally pooled and passed to a linear classifier with ten outputs.

For CIFAR-10, we use the same pooling configurations with a wider backbone. The convolutional channel widths are $3\to48\to64\to96\to128\to128$. The first four convolutional blocks include $2\times2$ stride-two pooling and reduce the spatial resolution from $32\times32$ to $2\times2$; the final convolutional block performs feature refinement without further downsampling. The resulting 128-dimensional representation is globally pooled and mapped to ten class logits by a linear layer.

\section{Lipophilicity regression on the original label scale}\label{app:lipo}

\begin{figure}[b!]
\centering
\includegraphics[width=0.6\textwidth]{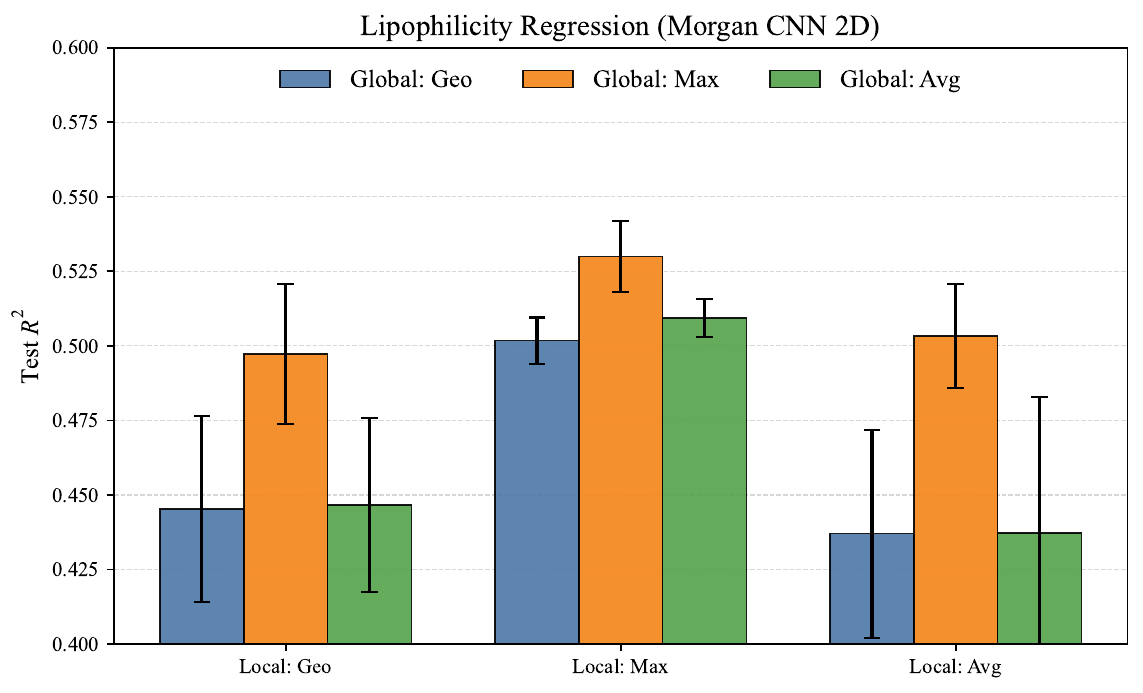}
\caption{
Test $R^2$ for lipophilicity regression on the original label scale using CNNs with different combinations of local and global pooling. Error bars denote the standard deviation across ten random seeds. Max pooling achieves the highest $R^2$ in this setting.
}
\label{sifig:lipolog}
\end{figure}

Figure~\ref{sifig:lipolog} presents the CNN pooling comparison on the original log-label scale. Max pooling gives the strongest performance in this setting, while the auxiliary exponential-target experiment favors a different configuration. This shift illustrates how target parameterization and its associated error weighting interact with the choice of pooling. In both settings, Softplus features allow the geometric-magnitude component to be examined within the same convolutional architecture.


\end{document}